\documentclass{article}

\PassOptionsToPackage{numbers, sort&compress, comma, square}{natbib}
\usepackage[preprint]{neurips_2023}

\usepackage[utf8]{inputenc} 
\usepackage[T1]{fontenc}    
\usepackage{url}            
\usepackage{booktabs}       
\usepackage{amsfonts}       
\usepackage{nicefrac}       
\usepackage{microtype}      
\usepackage{multirow}
\usepackage{caption}
\usepackage{subcaption}
\usepackage{graphicx}
\usepackage{xcolor}
\usepackage{listings}
\usepackage{amsmath}
\usepackage{enumitem}
\definecolor{lz}{rgb}{.224,.451,.686}
\usepackage[colorlinks, linkcolor=red, anchorcolor=blue, citecolor=lz]{hyperref}
\usepackage[normalem]{ulem}

\newcommand{\autoknow}{\textsc{SelfEvolve}}

\usepackage{xcolor} 
\definecolor{mygreen}{rgb}{0,0.6,0}
\definecolor{mygray}{rgb}{0.5,0.5,0.5}
\definecolor{mymauve}{rgb}{0.58,0,0.82}

\title{\autoknow: A Code Evolution Framework via Large Language Models}

\author{%
Shuyang Jiang$^1$, Yuhao Wang$^1$, Yu Wang$^{1,2 *}$ \\
  $^1$Shanghai Jiao Tong University\\
  $^2$Shanghai AI Laboratory \\
  \texttt{\{jiangshuyang,colane,yuwangsjtu\}@sjtu.edu.cn} \\
}
\begin{document}
\def\thefootnote{*}\footnotetext{Corresponding author}\def\thefootnote{\arabic{footnote}}

\maketitle

\begin{abstract}

Large language models (LLMs) have already revolutionized code generation, after being pretrained on publicly available code data. However, while various methods have been proposed to augment LLMs with retrieved knowledge and enhance the quality of code generation, the performance of these retrieval-based methods is limited by the strength of the retrievers used. In addition, while LLMs show great emergent ability, they still struggle to produce the correct code in one turn.
To address these challenges, we propose a novel two-step pipeline, called \autoknow, that leverages LLMs as both knowledge providers and self-reflective programmers. Unlike retrieval-based methods, \autoknow~obtains the knowledge from input prompts and generates intermediate code based on the generated knowledge. After that, \autoknow~asks LLM to act as an expert programmer to perform debugging for the generated code. This is achieved by receiving the error message from the interpreter, without requiring special test cases for correctness verification. 
We evaluate \autoknow~on three code generation datasets, including DS-1000 for data science code, HumanEval for software engineering code, and TransCoder for C++-to-Python translation. Our empirical experiments show that \autoknow~outperforms strong baselines by a significant margin on all datasets. We also conduct exhaustive analytical experiments to validate the effectiveness of the two stages of \autoknow, and find that both are superior to other prompting-based methods. Further scalability analysis demonstrates that \autoknow~can be adapted to other more advanced models, such as GPT-4, and bring consistent efficacy improvement.

\end{abstract}

\section{Introduction}

Code generation functions as a crucial and challenging component of various applications~\cite{hendrycks2021measuring,austin2021program,huang2022execution,lai2022ds}. 
However, the performance of large language models~(LLM) on diverse tasks and domains has substantially improved as the pretraining corpus expands. 
As a result, LLM has become the preferred model for code generation~\cite{chen2021evaluating,li2022competition}.
In fact, LLM performs much better than previous deep neural models dedicated to generating code~\cite{nadiel2022incoder,nijkamp2023codegen,lai2022ds}. 
Meanwhile, previous methods have been augmented by LLM's ability to digest various prompt contents and perform text generation~\cite{borgeaud2022improving,izacard2022atlas,mialon2023augmented}.
Various auxiliary augmentation signals have been added to the prompt to obtain more accurate code~\cite{shi-etal-2022-natural,bei2022codet,haluptzok2023language,zhou2023docprompting}.
However, most prior work usually obtains such signals via an external retriever and a large knowledge base. 
They leverage the problem description or natural language intents to retrieve relevant knowledge, including similar code snippets~\cite{parvez-etal-2021-retrieval-augmented,pasupat-etal-2021-controllable}, API documentation~\cite{xu-etal-2020-incorporating,zhou2023docprompting}, or focal methods~\cite{jin2023inferfix}.
Despite their success, retriever models can suffer from domain mismatch when adapting to different tasks, requiring finetuning or even training from scratch on the target domain, which limits their generality. 
Moreover, current retrievers are not well-suited for semi-structured knowledge items like library documentation, which can result in poor retrieved results.

To avoid domain mismatch and inaccurate retrieval results, we propose a two-stage paradigm, \autoknow, which treats LLM itself as a knowledge source. 
Previous work has demonstrated that LLM has encoded various domain knowledge~\cite{geva2020transformer} and can be treated as a large knowledge base~\cite{petroni-etal-2019-language,alkhamissi2022review}. 
Therefore, \autoknow~chooses to prompt LLM to generate multi-form necessary knowledge by itself. 
In particular, we prompt the language models to extract necessary knowledge from trial solutions or problem intents (\S\ref{self_generate}), depending on whether the problem intents contain explicit demands. 
This process excludes the intervention of retrievers since generating concrete knowledge based on roughly encoded ones in LLM parameters is easier than searching them in a large database with vague natural language statements. 
To the best of our knowledge, \autoknow~is the first LLM-driven self-augmented code generation framework.
Furthermore, inspired by the fact that human programmers rely on both related knowledge and a debugger to ensure implementation correctness, we inject an automatic refinement mechanism. 
This refinement mechanism teaches language models to depend on an executor like a Python interpreter to correct the preliminary code. 
We construct a runnable program from generated code and test cases extracted from the problem description (\S\ref{self_revise}) and execute it to obtain either pass or error messages, which serve as correction feedback. 
Compared to prompting LLM to generate test cases like CodeT~\cite{bei2022codet}, which may produce incorrect samples, \autoknow~maintains the correctness of test cases. 
Additionally, we do not grab evaluation samples from the test set like the recently proposed Self-Debugging~\cite{chen2023teaching}, which hardly generalizes to daily coding scenarios. 
Instead, the example cases in the problem description appear in coding tasks mostly and describe the behavior that the programmer needs to accomplish with little ambiguity. 
Leveraging these authentic and common test cases makes \autoknow~a reliable and general method for self-augmented code generation.

We primarily build \autoknow~using \texttt{gpt-3.5-turbo}~\footnote{\url{https://platform.openai.com/docs/api-reference/chat}}~(ChatGPT) and evaluate its performance on various tasks, including the data science code generation task DS-1000~\cite{lai2022ds}, the general code generation task HumanEval~\cite{chen2021evaluating}, and the C++-to-Python translation task TransCoder~\cite{roziere2020unsupervised}. 
Extensive experiments show that \autoknow~achieves a significant improvement in execution-based measurements compared to strong DocPrompting~\cite{zhou2023docprompting} and Self-Debugging~\cite{chen2023teaching} baselines on the data science code generation task~(\S\ref{ds_code_gen}). 
On HumanEval, \autoknow~still outperforms each strong baseline, and the self-refinement module brings noticeable performance improvements for the base LLM after using self-generated knowledge~(\S\ref{alg_code_gen_results}). 
Even on the code translation, which is a much simpler task for ChatGPT, \autoknow~still brings considerable improvement~(\S\ref{transcoder}). 
Furthermore, our analysis studies indicate that \autoknow~can provide more accurate knowledge than retrieval-based methods~(\S\ref{ablation_generate_know}), generalize to various datasets with only a few debugging turns~(\S\ref{ablation_iteration}), and scales easily to more powerful models like GPT-4~(\S\ref{scalability}). 
Finally, we use two intuitive cases to demonstrate how the two stages of \autoknow improve the generated code.
These cases illustrate the effectiveness of the model in generating high-quality code and highlight its potential for a range of applications.

\section{Related Work}

\paragraph{Augmented code generation}

In addition to problem descriptions and code snippets, many works provide auxiliary information to generate code. Before the era of LLM, researchers trained an encoder-decoder model that aimed to generate code based on a programmatic environment and function documentation~\cite{iyer-etal-2018-mapping}. JuPyT5~\cite{chandel2022training} conditions on Jupyter notebooks' context cells to generate data science code. Recently, large language models~(LLM), pretrained on a variety of corpus, have enabled an in-context learning pipeline for zero-shot or few-shot generation.
\citet{patrick2022language} introduced a method to solve programming puzzles with synthetic puzzles and solutions generated by LLM. \citet{parvez-etal-2021-retrieval-augmented} augmented code generation models with retrieved similar code snippets. In contrast, API-specific documents retrieved via a CodeT5~\cite{wang-etal-2021-codet5} retriever serve as additional information in the prompt~\cite{zhou2023docprompting}. However, their methods involve retrieving semi-structured knowledge items, whose performance is bottlenecked by current retriever models. Moreover, it is hard for retrievers to adapt to the target domain when the corpus for finetuning the retriever is inaccessible.
Compared to theirs, LLM is more suitable for bridging the gap between domains than any small retriever, which provides more accurate knowledge. Moreover, our method does not require domain-specific finetuning, offering higher accessibility and generality.

\paragraph{Automatic code refinement}

Language models often output unreliable information with high confidence~\cite{ji2023survey}. This unreliability is reflected in bug code snippets when generating code. Some previous works have trained exclusive models dedicated to repairing incorrect code, which only accept the bug code as input~\cite{gupta2017deepfix,yasunaga2021break}. Other works have fused auxiliary information obtained from execution, such as the stack trace~\cite{gupta2020synthesize,wangdynamic} and error information raised by a compiler~\cite{yasunaga2020graph}. In recent work, LLMs have been leveraged to act as ``teachers'' to fix bugs hidden in the code. 
CodeT~\cite{bei2022codet} uses synthetic test cases obtained through Codex~\cite{chen2021evaluating} to select correct codes. However, CodeT is applied in an unnatural scenario where all coding problems are formatted as writing a function with certain inputs. 
\citet{madaan2023self} iteratively refines the output with model-generated feedback in multiple tasks, except for code generation. 
\citet{chen2023teaching} uses similar methods in code generation, but their work peeps one test case from the ground truth, which is impossible when solving real problems.
Without exposing test cases, our method is much more flexible and closer to a real coding scenario.

\begin{figure}[tbp]
  \centering
  \includegraphics[width=\linewidth]{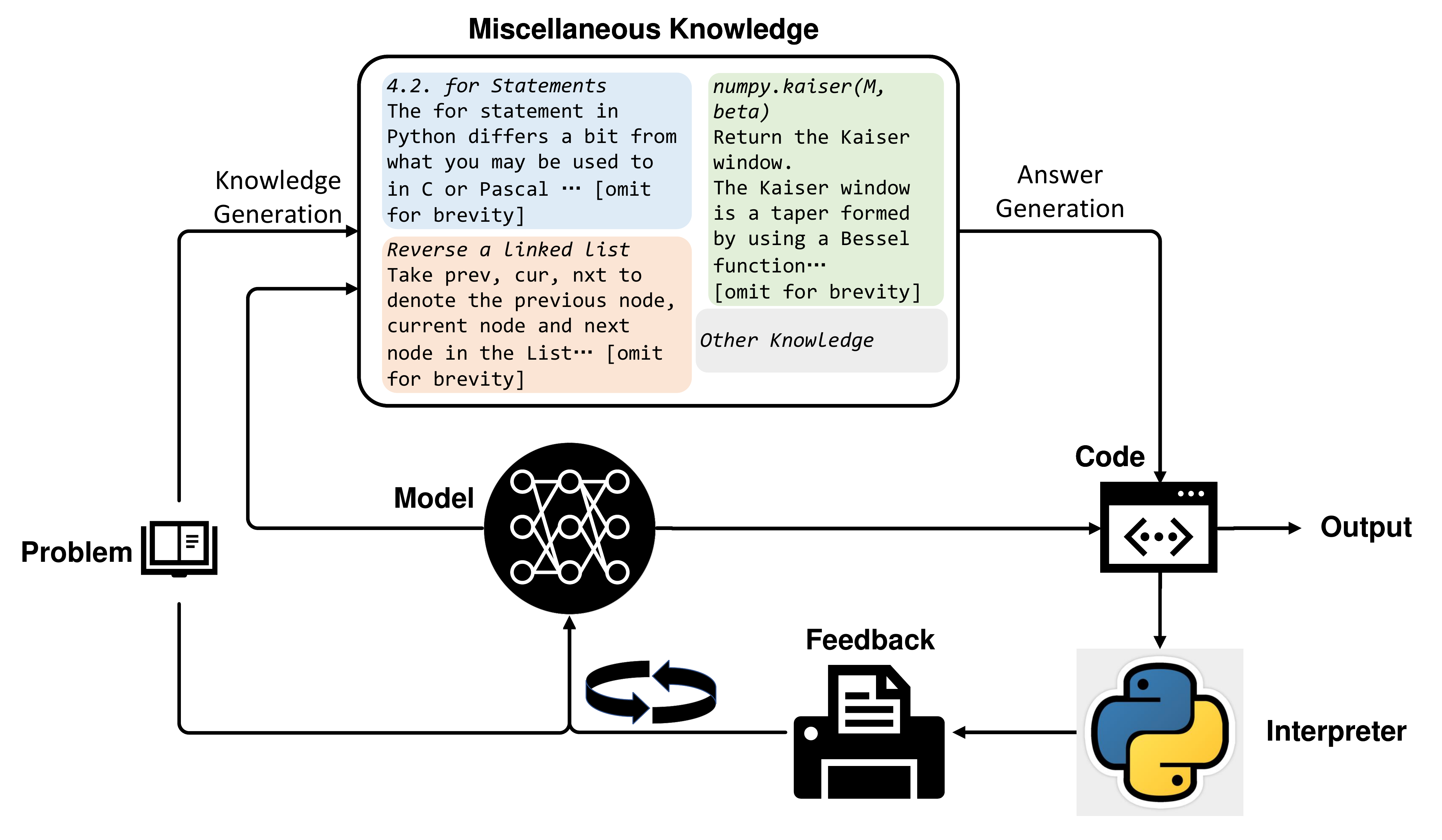} 
  \caption{The \autoknow~pipeline. LLMs first generate corresponding knowledge for the related problem, then generate the trial answer conditioned on the knowledge. The iterative refinement step uses test cases and generated code snippets to form executable programs and then prompts LLM to refine the answer code based on the feedback thrown by the interpreter. }
 \label{fig:selfevolve}
\end{figure}

\section{\autoknow: Code Evolution via Large Language Models}
\label{autoknow_overview}

This section first briefly introduces the code generation paradigm inspired by natural programming practice. We then present the concept of \autoknow, a two-step method that utilizes language models as a self-bootstrapping knowledge enhancer and an expert programmer with self-reflection, without external models or knowledge bases.
After presenting the concept, we delve into the two primary components of \autoknow, which are built up to form a fully LLM-driven generation pipeline without the need for fine-grained prompt designs or further finetuning steps. The overall pipeline of our method is presented in Figure~\ref{fig:selfevolve}.

\subsection{BackGround}

\paragraph{Code generation formulation}
Given a problem description written in natural language $d$, and code context $c$, the autoregressive language model $p_{\theta}$ predicts the solution as 
\begin{equation}
    P(Y)=\prod_{i=1..n} p_{\theta}(Y_i|Y_{<i}, X), Y_{<1}=\emptyset
\end{equation}
where $n$ is the prediction length and $X=[d;c]$ is the concatenation of $d$ and $c$.

\paragraph{Two-step code generation pipeline}
Conditioning solely on problem description for generation is still hard for LLM.
Inspired by most programmers who often refer to knowledge documentation~\cite{roehm2012professional} and struggle to debug with current tools~\cite{parnin2011automated}, we divide prompt-based code generation methods into two steps.
The first step prompts language models to comprehend extra knowledge and task-specific instructions~\cite{brown2020language,izacard2022few,mialon2023augmented,peng2023check}, while the next one teaches models to revise the generated code solution through feedback from humans or an oracle instructor.
In the two-step pipeline, the second generation step never deteriorates the intermediate output of the first step.
Therefore, these two steps follow a topological order in terms of optimization, and can be optimized in order and fused together.

\subsection{\autoknow: the Proposed Two-step Pipeline}
\label{autoknow_detail}

Based on the above analysis, we propose \autoknow, which improves both steps by enabling generated code to evolve progressively using only a large language model, without requiring any learning.
\autoknow~generates code by conditioning on the knowledge in the prompt, as previous work has done. However, the knowledge is generated by LLM instead of being retrieved from external knowledge bases.
After obtaining the output of the first step, \autoknow~uses LLM to iteratively revise the generated code. 
This process follows~\citet{chen2023teaching} to correct code errors by leveraging feedback from a code executor, but does not necessitate the use of specific test cases.

\paragraph{Generating knowledge with language models}
\label{self_generate}
Conditioning a language model on the knowledge in the prompt is crucial, yet challenging. Given $m$ knowledge items $K[1..m]$, the language model predicts the next token to generate the final code solution:
\begin{equation}
    P(Y|K)=\prod_{i=1..n} p_{\theta}(Y_i|Y_{<i}, X, K), Y_{<1}=\emptyset
\end{equation}
Knowledge can be retrieved via a sparse retriever~\cite{robertson1994some} or a dense retriever~\cite{reimers-2019-sentence-bert,gao-etal-2021-simcse} as such:
\begin{equation}
    K :=\arg\max_{K\subset B} P(K| X, B)
\end{equation}
where $B$ is the whole database.
However, the performance of current retriever models may be limited, resulting in $K$ containing irrelevant knowledge items that add noise to LLM and harm the generation results.
A widely-used approach to mitigate this problem is to retrieve as much knowledge as possible~\cite{borgeaud2022improving} to cover the necessary items. However, this method places demands on the ability of LLM to process long texts, which is still a work in progress~\cite{PENG_RWKV-LM_2021,li2023context}.

To more accurately and conveniently obtain the necessary knowledge, we utilize language models as knowledge sources, prompting them to generate information. Large language models have encoded knowledge from a variety of databases into their parameters after being pre-trained on various corpora~\cite{geva2020transformer}. Additionally, models that undergo reinforcement learning from human feedback (RLHF)~\cite{ouyang2022training} can follow human instructions, serving as a natural knowledge source and providing miscellaneous knowledge based on appropriate input instructions.
Based on this, we propose to use self-generated knowledge which is fetched via prompting LLMs as such:
\begin{equation}
\label{eq:gen_know_direct}
    p(K) = \prod_{i=1..k} p_{\theta}(K_i | X, K_{<i}), K_{<1}=\emptyset
\end{equation}
where $k$ is the length of generated knowledge tokens.
When problem descriptions contain implicit intents, such as in StackOverflow~\cite{lai2022ds}, there is often a gap between the detailed knowledge required and the words used to describe the problem. This gap arises because deriving the required knowledge involves reasoning.
To narrow this reasoning gap and obtain more precise knowledge, we decompose this extraction process when intents are implicitly given:
\begin{equation}
    \label{eq:gen_know_2hop}
    p(K) = \prod_{i=1..k} p_{\theta}(K_i | c, K_{<i})\cdot p_{\theta}(c|X), K_{<1}=\emptyset
\end{equation}
where $c$ is a trial code solution from LLM based only on problem contexts.
It contains necessary but potentially misused knowledge that can benefit the extraction of knowledge ($K$). $K$ can be formatted as any problem-specific structure to fit the problem instruction $X$, making it suitable for various tasks. When problem descriptions $X$ contain explicit intents, \autoknow~uses Eq.~\ref{eq:gen_know_direct} instead, as LLM can easily extract knowledge with high accuracy in this case.

\paragraph{Revision of generated solution}
\label{self_revise}
Previous studies have shown that intermediate results generated from LLM may contain mistakes~\cite{wei2022chain,zhang2022automatic,wang2022self,lyu2023faithful}.
Such errors can introduce noise to the prompt context, reducing the accuracy of the final output.
To reduce code errors, we mimic the debugging process of programmers and introduce an iterative self-refinement mechanism to rectify the buggy code.
This mechanism leverages an external Python interpreter to revise erroneous programs.
Our approach incorporates code context and sample test cases into the input prompt, along with the generated code solution, to form an executable program. We then execute this program in a sandbox environment to receive error information as well as standard output. Once error information is obtained, we prompt language models to revise the buggy programs, conditioned both on program requirements and error information:
\begin{equation}
    P(Y'|X, Y, K, e) = p_{\theta}(Y'| X, Y, e)\cdot p_{\theta}(Y | X, K)
\end{equation}
The revised output, $Y'$, may still contain bugs. Therefore, the above process is repeated until the code can be interpreted without exceptions, or until the iteration steps reach a fixed threshold.
In practical applications, the modeling of $p_{\theta}(Y'|X, Y, e)$ varies depending on the type of error. For simplicity, \autoknow~only corrects API errors and incorrect assertion statements. We find that correcting these two types of errors contributes significantly to performance improvement in the empirical experiments.

In conclusion, we combine two LLM-driven methods - generation based on self-generated knowledge and refinement via error message - to create a more effective method called \autoknow. These two components reinforce each other almost orthogonally.
On one hand, self-generated knowledge boosts self-refinement steps. By conditioning on knowledge from models' parameters, intermediate output explicitly applies the knowledge. With more accurate output, the self-refinement steps require fewer iterations to repair the code, resulting in lower difficulty.
On the other hand, the self-refinement steps improve the application of generated knowledge. The input knowledge may contain irrelevant information or noise during the generation process. The self-refinement steps eliminate this noise by introducing an external interpreter, improving the overall quality of the generated code.
Later empirical experiments will demonstrate how these two modules reinforce each other.

\section{Experiments}

In this work, we present a novel pipeline that supports natural and reliable code generation for a variety of programming and data science problems. To evaluate its effectiveness, we conducted experiments using three different code generation tasks: data science code generation, simple algorithm coding, and C++-to-Python code translation. These tasks were assessed using the DS-1000~\cite{lai2022ds}, HumanEval~\cite{chen2021evaluating}, and TransCoder~\cite{roziere2020unsupervised} benchmarks, respectively. In all experiments, we set the top-p cutoff to 0.95 and the maximum generation length to 1024. For the specific prompts used for each task, please refer to Appendix~\ref{sec:each_task_gen_prompt} and~\ref{sec:each_task_refine_prompt}.

\subsection{Baselines}
\begin{enumerate}[itemsep=0.8mm, parsep=0pt, leftmargin=*]
 \item \textbf{DocPrompting}~\cite{zhou2023docprompting}: DocPrompting improves the LLM by retrieving problem-relevant documentation via a finetuned retriever, then condition on those documents and problem description to generate code. We use the same documentation pool for DocPrompting in DS-1000, as the problem source of DS-1000 is the same as that of CoNaLa~\cite{yin2018mining}. We also use the same retrieval weights released by them, as DS-1000 are also built to test Python programming.
 \item \textbf{Self-Debugging}~\cite{chen2023teaching} relies on a SQL application or Python interpreter to teach language models to revise SQL commands or Python code with bugs. They propose three debugging ways, including ``simple'', ``unit test'' and ``explanation''. Without training sets in DS-1000 and HumanEval, we implement it in a zero-shot way and use its ``simple'' variant for a fair comparison.
 \item \textbf{\autoknow}: \autoknow~is the code generation pipeline proposed in this work. In the main experiments, We use ChatGPT as the knowledge generator and the code refiner. 

\end{enumerate}

\subsection{Main Results of~\autoknow}

\begin{table}[hbtp]
  \centering
  \caption{Pass@1 results on the DS-1000 dataset. $^\dag$ denotes that the results are referred from~\cite{lai2022ds}. Other baselines are implemented with the same prompt and hyperparameter setting.}
    \begin{tabular}{lccccc}
    \toprule
    \multicolumn{1}{c}{\multirow{2}[4]{*}{\textbf{Method}}} & \multicolumn{4}{c}{\textbf{Perturbation}} & \multicolumn{1}{c}{\multirow{2}[1]{*}{\textbf{Overall}}} \\
\cmidrule{2-5}          & Origin & Surface & Semantic & Diff-Rewrite  \\
    \midrule
    \multicolumn{6}{c}{\textit{Prior work}} \\
    \midrule
    Codex (Completion)$^\dag$ & 44.93 & 37.94 & 34.35 & 16.94 & 39.20 \\
    Codex (Insertion)$^\dag$ & 47.76 & 50.18 & 38.39 & 21.05 & 43.30 \\
    DocPrompting & 53.95 & 50.00 & 39.57 & 25.93 & 45.50 \\
    Self-Debugging & 63.38 & 59.21 & 45.65 & 28.40 & 53.00 \\
    \midrule
    \multicolumn{6}{c}{\textit{This work}} \\
    \midrule
    ChatGPT & 60.31 & 52.63 & 41.30 & 26.54 & 49.30 \\
    \autoknow  & \textbf{66.23} & \textbf{67.11} & \textbf{48.70} & \textbf{33.95} & \textbf{57.10} \\
    \qquad \textit{w/o self-refinement} & 60.09 & 59.21 & 41.30 & 29.01 & 50.60 \\
    \bottomrule
    \end{tabular}%
  \label{tab:ds1000}%
\end{table}%

\paragraph{Data science code generation}
\label{ds_code_gen}

For data science code generation tasks, We selected the DS-1000~\cite{lai2022ds} benchmark, which contains 1000 problems covering seven common data science libraries. DS-1000 introduces a novel ``perturbation'' concept, including Origin, Surface, Semantic, and Diff-Rewrite, representing the difficulty of problems in ascending order, making it a challenging benchmark.
In this study, we prompted language models to generate problem-relevant API documentation as domain-required knowledge. For the self-refinement module, we checked the executable programs and prompted language models to fix syntax errors only. We used greedy decoding and reported the pass@1~\cite{chen2021evaluating} score for each method. Results are presented in Table~\ref{tab:ds1000}.
Without further refinement steps, \autoknow~has already exceeded the strong ChatGPT baseline on the Surface and Diff-Rewrite perturbation types, by a margin of 6.58 and 2.47, respectively. Moreover, with an additional self-debug module, \autoknow~substantially improves over other baselines, with a 7.8~(\textit{relatively 15.8\%}) pass@1 gain compared to ChatGPT, on average. \autoknow~also surpasses the prompt-based method, Self-Debugging, by a convincing 4.1 performance margin.
We also noticed that integrating the self-generated knowledge with the self-refinement module gains much higher improvement. 
Specifically, \autoknow~improves the baseline in terms of all perturbation types, demonstrating that our method can impressively enhance the robustness of large language models.

\paragraph{General code generation}
\label{alg_code_gen_results}

For general code generation tasks, we evaluated \autoknow~on HumanEval~\cite{chen2021evaluating}. This dataset contains 164 hand-written Python programming problems with an average of 7.7 test cases each. We implemented Self-Debugging methods on ChatGPT and reported its score. We did not implement DocPrompting since no library documentation is required in HumanEval. We also introduced the GPT-4 results from~\cite{bubeck2023sparks} for comparison.
We reported a pass@1 score for greedy decoding and pass@10 for 10-sample decoding. For the 10-sample generation, we conducted a grid search to set the temperature to $t=1$. Unlike DS-1000, we induced LLM to explicitly output problem-related algorithms as external knowledge and taught LLM to fix assertion errors and syntax errors.
The results in Table~\ref{tab:HumanEval} demonstrate that the strong ChatGPT baseline significantly benefits from our \autoknow~method, with an 11.59 pass@1 gain and a 6.71 pass@10 gain. This leaves a small 3.95 pass@1 gap from GPT-4. Notably, with self-generated knowledge, the self-refinement module again harvested a larger improvement (+3.66 pass@10) than only applying a refinement module like Self-Debugging (+1.22 pass@10). This empirically verifies that self-generated knowledge helps to reduce most errors and produce more precise results.

\begin{figure}[tbp]
\begin{minipage}[t]{0.48\textwidth}
\centering
\captionof{table}{{Pass@1 and pass@10 scores comparisons with different methods on HumanEval. We use the same prompt to implement each method. $^\dag$ denotes that scores are cited from~\cite{bubeck2023sparks}. 
}
}
\label{tab:HumanEval}
\resizebox{1.0\columnwidth}{!}{%
    \begin{tabular}{lcc}
    \toprule
    \textbf{Model}   & \multicolumn{1}{l}{\textbf{Pass@1}} & \multicolumn{1}{l}{\textbf{Pass@10}} \\
    \midrule
    \multicolumn{3}{c}{\textit{Prior Work}} \\
    \midrule
    GPT-4$^\dag$     & 82.00  & - \\
    \texttt{text-davinci-003}$^\dag$      & 65.00 & - \\
    ChatGPT      & 66.46  & 86.58 \\
    CodeT~\cite{bei2022codet}     & 65.20 & 86.80 \\
     Self-Debugging         & 73.78  & 87.80  \\
    \midrule
    \multicolumn{3}{c}{\textit{Ours}} \\
    \midrule
        \autoknow      & \textbf{78.05 } & \textbf{93.29 } \\
    \qquad \textit{w/o self-refinement}     & 70.73  & 89.63  \\
    \bottomrule
    \end{tabular}%
}
\vspace{-1em}
\end{minipage}
\hfill
\begin{minipage}[t]{0.48\textwidth}
\centering
\captionof{table}{Performance comparison on TransCoder dataset where we follow~\cite{roziere2020unsupervised,chen2023teaching} to translate C++ code to Python code. All methods in this work are implemented with greedy decode. ``Acc.'' refers to computational accuracy. }
\label{tab:transcoder}
\resizebox{0.9\columnwidth}{!}{%
\setlength\tabcolsep{9pt}
\begin{tabular}{lcc}
    \toprule
    \multirow{2}[4]{*}{\textbf{Method}} & \multicolumn{2}{c}{\textbf{TransCoder}} \\
\cmidrule{2-3}          & \textbf{Acc.}  & \textbf{Pass@1} \\
    \midrule
    \multicolumn{3}{c}{\textit{Piror Work}} \\
    \midrule
    PaLM~\cite{chowdhery2022palm} & 51.8  & - \\
    PaLM-Coder~\cite{chowdhery2022palm} & 55.1  & - \\
    Codex~\cite{chen2021evaluating} & 80.4  & - \\
    Self-Debugging~\cite{chen2023teaching} & 89.3  & - \\
    \midrule
    \multicolumn{3}{c}{\textit{Ours}} \\
    \midrule
    ChatGPT & 92.7  & 90.0 \\
    \autoknow~ & \textbf{94.8} & \textbf{92.4} \\
    \qquad \textit{w/o self-refinement} & 93.4  & 90.5 \\
    \bottomrule
    \end{tabular}%
}
\end{minipage}
\end{figure}

\paragraph{Python code translation}
\label{transcoder}

As suggested by \citet{roziere2020unsupervised}, we experimented with our methods on the TransCoder~\cite{roziere2020unsupervised} dataset. We used its test set, which requires translating C++ code to Python, and filtered out problems without testing scripts, resulting in 410 valid problems.
In addition to pass@1, we followed~\citet{roziere2020unsupervised} by using another evaluation metric, computational accuracy, to test each model. Computational accuracy computes the accuracy score for each problem in a competition rule, where each sample code is scored as the percentile of its passed test cases, while the pass@1 metric is computed as whether the sample code has passed all test cases.
We prompted LLM to generate the algorithm detail of the C++ code, which serves as the context for Python code generation. Results in Table~\ref{tab:transcoder} indicate that our proposed method, \autoknow, achieves the best performance among other prompt-based methods, even outperforming Self-Debugging which peeps one ground truth test case. Built upon a strong ChatGPT baseline, \autoknow~further improves by 2.1 computational accuracy and 2.4 pass@1, respectively. Without the self-refinement module, \autoknow~still improves over ChatGPT, with a 0.7 accuracy gain and 0.5 pass@1 gain.

\subsection{Discussion}
In this section, we conduct various analysis experiments to validate the efficacy of our proposed \autoknow~.
We first present the impact of the number of iteration step on the final performance of \autoknow.
After that, we demonstrate how our generated knowledge is superior to retrieved knowledge, through a human evaluation experiment.
Finally, we extend our method to an even more intelligent language model~(GPT-4) to empirically show the scalability of \autoknow.

\begin{figure*}[bp]

    \begin{subfigure}[c]{0.47\textwidth}
         \centering
         \includegraphics[width=\textwidth]{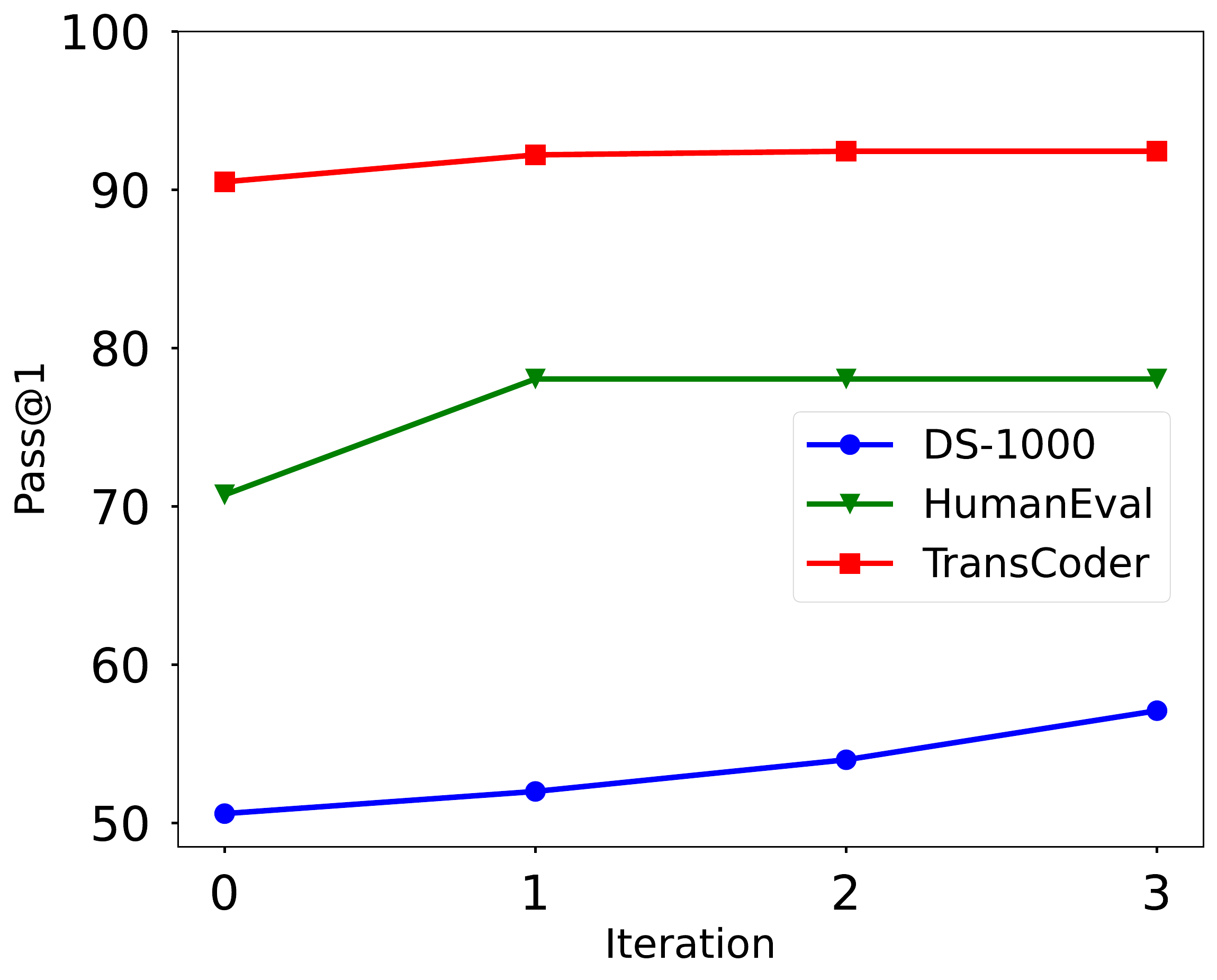}
         \caption{
         }
         \label{fig:iter_analysis}
     \end{subfigure}
     \hspace{\fill}
     \begin{subfigure}[c]{0.47\textwidth}
         \centering
         \includegraphics[width=\textwidth]{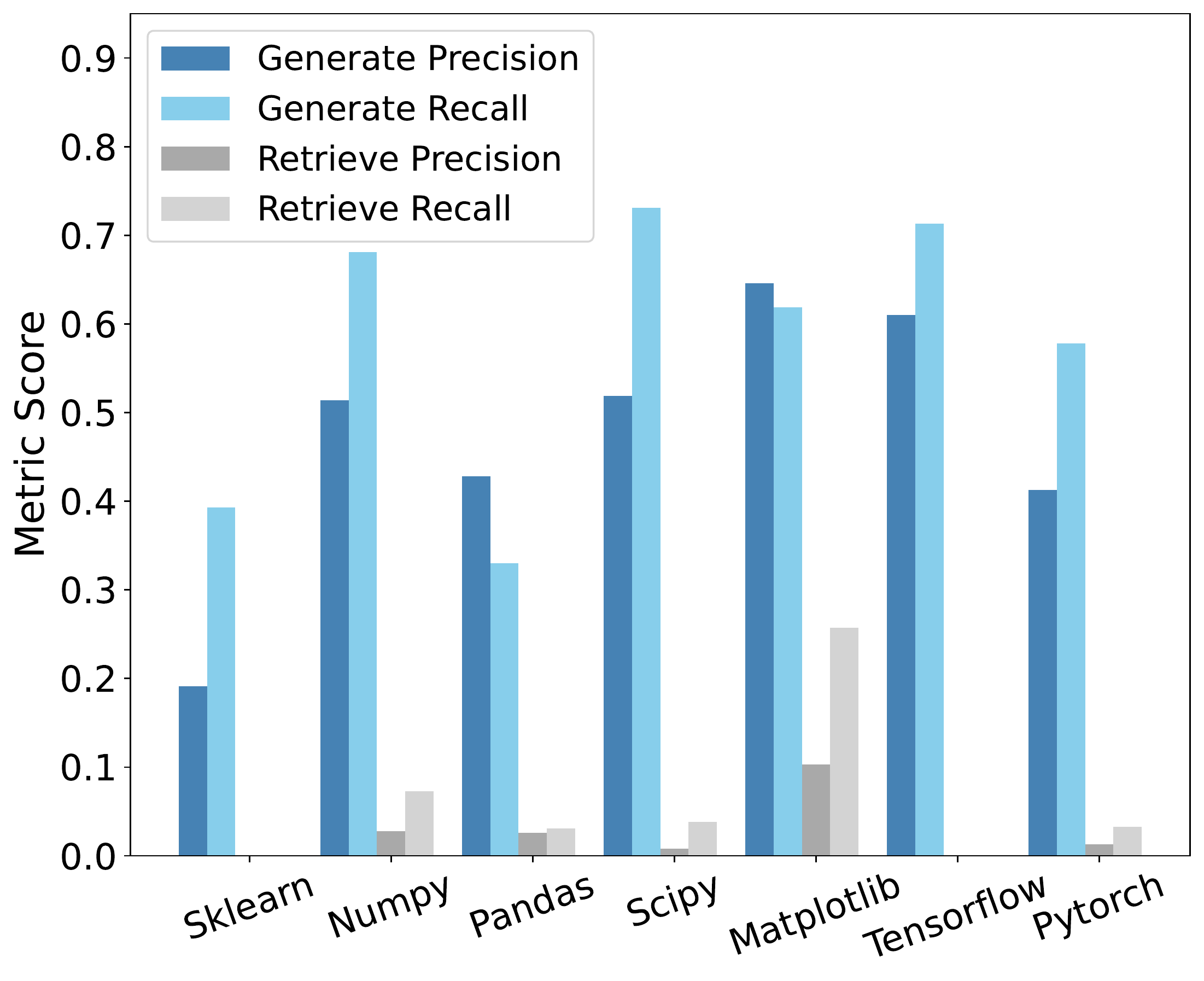}
         \caption{
         }
         \label{fig:acc_gen}
     \end{subfigure}

    \hspace*{\fill}
    \caption{(a) Performance-iteration curves of \autoknow~on DS-1000, HumanEval and TransCoder datasets. (b) Precision and recall comparisons between generated knowledge and retrieved one.}
    \label{fig:two_tradeoff}
\end{figure*}

\paragraph{How do iteration steps affect performance?}
\label{ablation_iteration}

In \S\ref{self_revise}, we declared that the refinement module is iteratively run to fix bugs. In this experiment, we determine under what conditions we should stop the refinement module. We ran the self-refinement module for different iteration turns on three datasets using greedy decoding and tested the pass@1 score of ChatGPT using the same prompt for each iteration stage.
Figure~\ref{fig:iter_analysis} presents the detailed results. We observed that the major improvement came from the first refinement step for the HumanEval and TransCoder datasets. On DS-1000, however, the performance improved almost uniformly as we increased the number of refinement steps until the third iteration. This discrepancy across datasets resulted from the much more difficult nature of the problems in DS-1000 compared to the other two datasets. Therefore, the self-refinement module brought consistent improvement under different refinement stages.
This finding suggests that \autoknow~requires more refinement steps when processing difficult problems, but only one debugging turn is sufficient to bring major improvement for less complicated problems.

\paragraph{Human evaluation of self-generated knowledge}
\label{ablation_generate_know}

To better understand the superiority of generated knowledge in realistic scenarios, we conducted a human evaluation study to demonstrate that the generated knowledge is more relevant to the problem topic than a retrieval-based one. 
We randomly selected 200 problems from seven libraries of DS-1000 and asked two data science experts to count the number of correctly provided API documents to determine whether the API knowledge matched the solution. 
We then used two common metrics, precision and recall~\cite{buckland1994relationship}, to assess the accuracy of the knowledge in accordance with the oracle answer. 
Precision is defined as the percentage of correctly provided documents in the provided document set, while recall measures the percentage of correct document items in the oracle document set. 
More details on the experiment are presented in Appendix~\ref{detail_of_human_evaluation}. 
The comparison bar chart is shown in Figure~\ref{fig:acc_gen}. 
We observed that in all libraries, the generated knowledge was much more accurate than the retrieved one in both metrics. Notably, the retrieved knowledge showed little match with oracle solutions in most libraries because the retrieval queries in DS-1000 are too complicated and contain implicit API demands. In the Matplotlib library, where the queries are simple and the demands are explicitly stated, the retrieved knowledge matched the problem requirements slightly but still lagged far behind the generated one. 
One key reason for the superiority of generated knowledge is that LLMs can bridge the reasoning gap between problem descriptions and knowledge terminology better than a retriever model. 
This is also the reason behind Eq.~\ref{eq:gen_know_2hop}. In other words, generated knowledge is able to provide a more comprehensive and accurate understanding of the problem topic, which is crucial in realistic scenarios.

\paragraph{Scaling to more powerful models}
\label{scalability}

To evaluate the scalability of \autoknow~in more advanced language models, we integrated \autoknow~with GPT-4 without requiring excessive prompt engineering. GPT-4 has demonstrated significantly greater intelligence and reasoning abilities compared to ChatGPT~\cite{bubeck2023sparks}. However, due to limited GPT-4 API-Key access, we only conducted experiments on the Scipy, Pytorch, Sklearn, and Matplotlib libraries of DS-1000, which includes a total of 444 problems and HumanEval.
We used the same prompt as ChatGPT and used greedy decoding to report the pass@1 score. The results for both datasets are shown in Table~\ref{tab:gpt4}. This experiment demonstrates that our approach can benefit from more advanced backbone models instead of degrading them. In contrast to the ChatGPT-based version, \autoknow~on GPT-4 achieved higher pass@1 scores on Scipy~(+4.72), Pytorch~(+19.12), and Sklearn~(+6.09). A similar performance gain was also observed in HumanEval, where \autoknow~improved the already higher pass@1 score by 2.76. Furthermore, after adding the lightweight self-refinement plugin, GPT-4 demonstrated further improvements on all datasets. This highlights the effectiveness of leveraging advanced backbone models and the potential of our approach in producing superior results.

\begin{table}[tbp]
  \centering
  \caption{Comparison between \autoknow~using ChatGPT and GPT-4 baselines. We bind \autoknow~with ChatGPT and GPT-4 to test its generalization.}
    \begin{tabular}{lccccc}
    \toprule
    \multirow{2}[4]{*}{\textbf{Method}} & \multicolumn{4}{c}{\textbf{DS-1000}} & \multirow{2}[4]{*}{\textbf{HumanEval}}\\
\cmidrule{2-5}          & Scipy & Pytorch & Sklearn & Matplotlib & \\
    \midrule
    \autoknow~(ChatGPT) & 52.83 & 64.71 & 73.04 & 78.06 & 78.05 \\
    GPT-4 & 52.83 & 44.12 & 60.00 & 69.03 & 82.00 \\
    \autoknow~(GPT-4) & \textbf{58.49} & \textbf{70.59} & \textbf{70.43} & \textbf{84.52} & \textbf{89.02}\\
    \qquad \textit{w/o self-refinement} & 57.55 & 63.24 & 66.09 & 69.03 & 84.76 \\
    \bottomrule
    \end{tabular}%
  \label{tab:gpt4}%
\end{table}%

\subsection{Case Study}

\begin{figure*}[tbp]

    \begin{subfigure}[c]{0.47\textwidth}
         \centering
         \includegraphics[width=\textwidth]{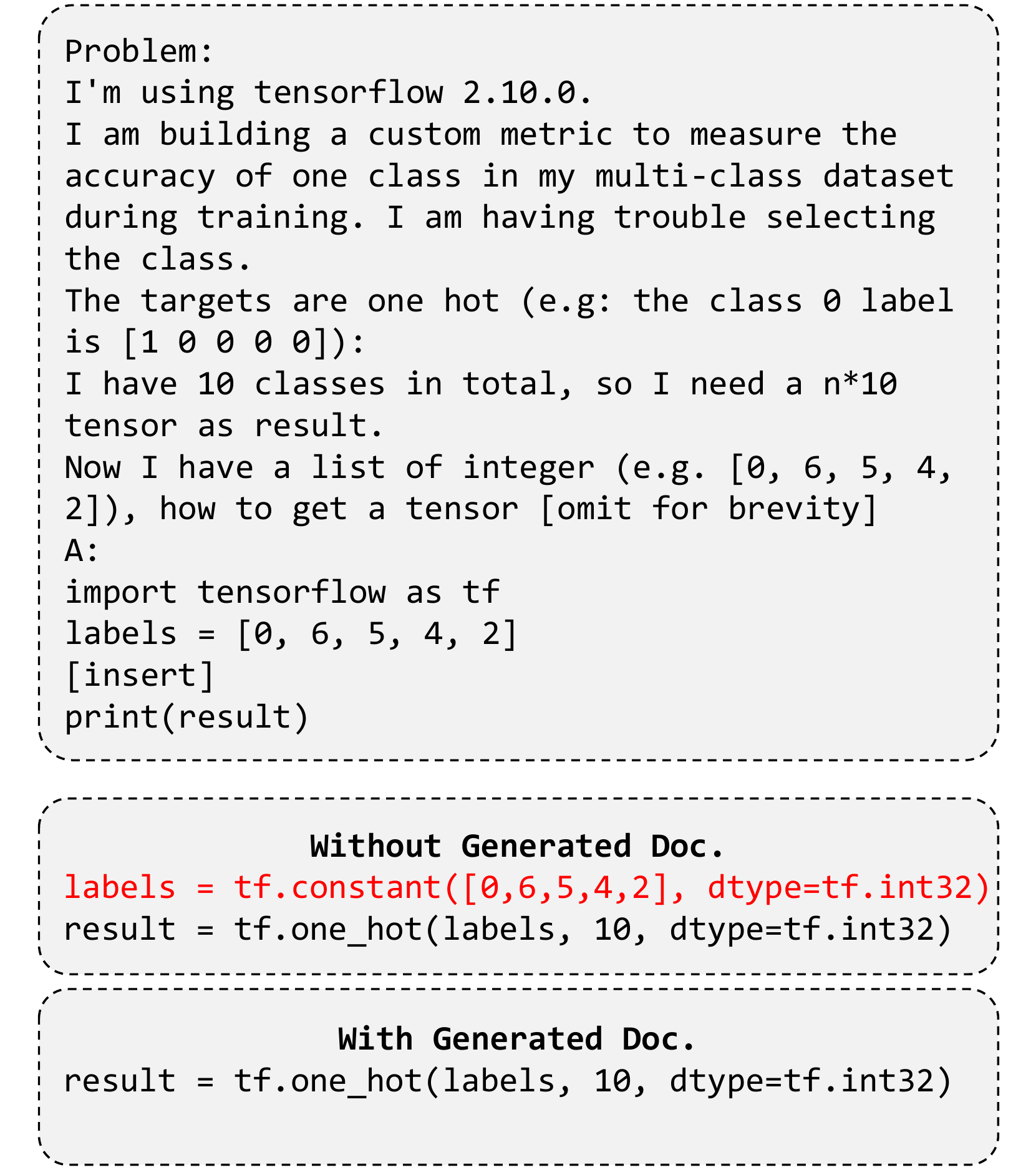}
         \caption{
         }
         \label{fig:doc_example}
     \end{subfigure}
     \hspace{\fill}
     \begin{subfigure}[c]{0.47\textwidth}
         \centering
         \includegraphics[width=\textwidth]{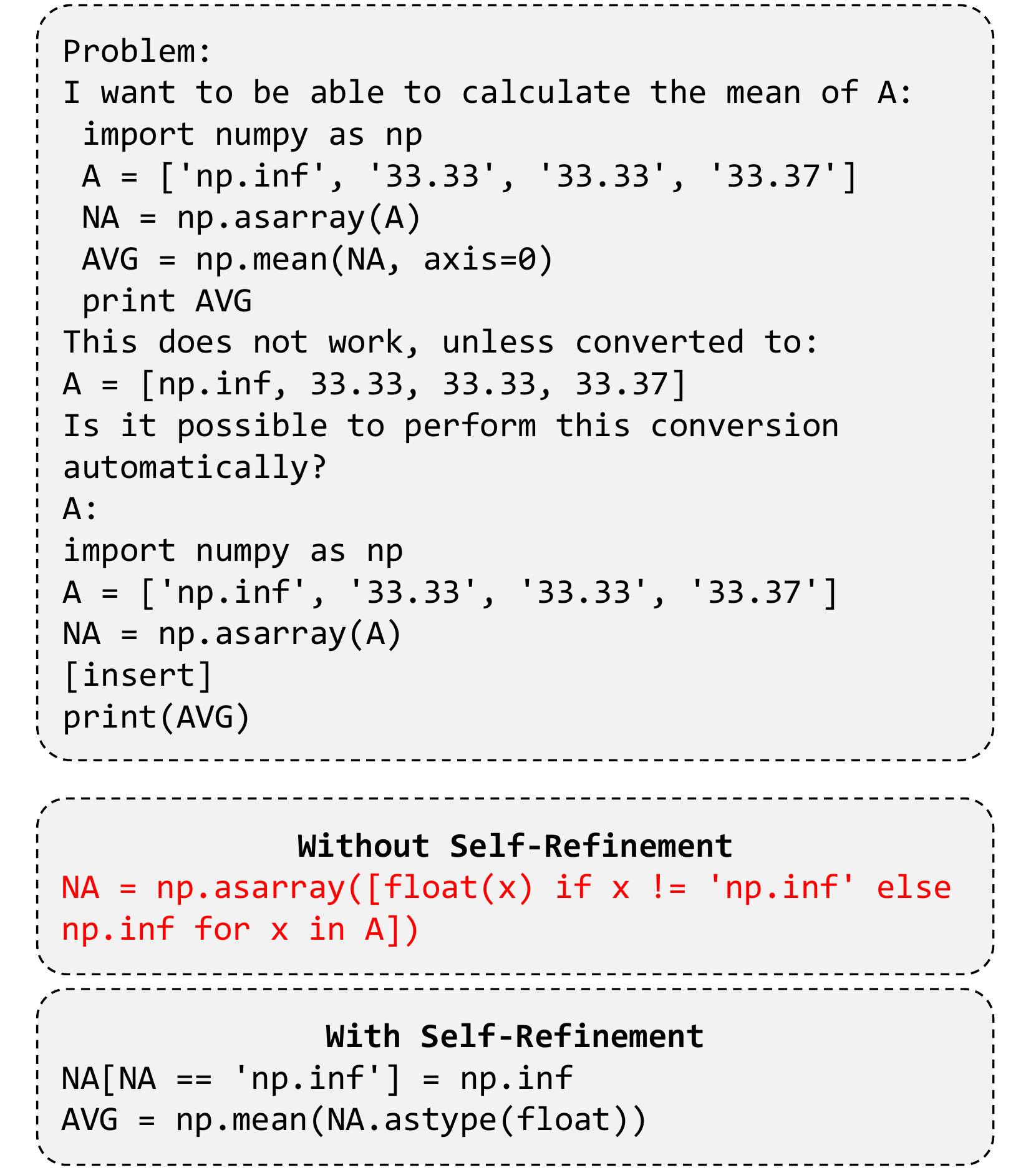}
         \caption{
         }
         \label{fig:revise_example}
     \end{subfigure}

    \hspace*{\fill}
    \caption{Two examples to show the efficacy of our proposed \autoknow~methods, where red codes are wrong codes. (a) Comparison between with and without generated documentation. (b) Comparison between with and without self-refinement module.}
    \label{fig:case_study}
\end{figure*}

This section demonstrates the effectiveness of \autoknow~with two representative examples shown in Figure~\ref{fig:case_study}. In the first example, LLM generates specific documentation \texttt{tf.one\_hot} for the problem. Compared to the vanilla output without documentation, language models output extra codes that cannot generalize to other test cases. In contrast, conditioning on the concrete API documentation, language models output more deterministic code without transforming the original labels to a tensor type. The provided documentation sharpens the probability curve and contributes to a more accurate and general answer.
In the second example, language models read the \texttt{np.asarray} documentation, but forgets to compute the \texttt{AVG} value. By leveraging the traceback information without the specific test cases, language models can revise their code and use a more general method to solve the problem.
Both examples illustrate how the two methods in \autoknow~enhance each other to improve performance. \autoknow~helps make the output of language models more general and accurate.

\section{Limitation \& Future Work}

Although \autoknow~has shown promising results in generating knowledge and improving the performance of large language models, there are still some limitations that need to be addressed. One of the main challenges of \autoknow~is that it may not always be automatic when used in different tasks due to the hand-written prompting words. This means that its effectiveness may be limited when applied to other use cases.
Another limitation of \autoknow~is that the generated knowledge may not always be suitable for every task and may require fine-grained selection to be effective. 
However, these issues can be mitigated by developing suitable prompting skills. For example, a more comprehensive set of prompting words could be developed to make it easier to adapt \autoknow~to new tasks. 
Additionally, a more sophisticated algorithm could be developed to automatically select the appropriate knowledge for a given task. 
We believe that addressing these issues will make \autoknow~a more versatile and useful framework in different contexts.

\section{Conclusion}

We propose \autoknow, a simple yet effective method for solving code generation problems using large language models (LLMs) as a fully LLM-driven framework. It acts as both a knowledge provider and a self-reflective programmer to generate high-quality code in two steps, both of which are run with a single LLM. This makes it more flexible and extendable to other datasets, such as Spider~\cite{yu-etal-2018-spider} or APPS~\cite{hendrycks2021measuring}, without requiring an extra retriever or previously set up database. Substantial experiments on diverse code generation tasks have verified that \autoknow~can bring great performance gains under various tasks and datasets, and outperform two strong prompting-based methods by a good margin. Furthermore, various analysis experiments indicate that \autoknow~is more suitable for providing problem-related knowledge compared to traditional dense retrievers, and can be easily scaled to more intelligent language models to bring further improvements.

{
\small
\bibliography{ref}
}


\appendix
\section{Comparison between Generated Knowledge and Retrieved Knowledge}
\label{detail_of_human_evaluation}
We here present the experiment details of this human evaluation experiment.
We randomly select one-fifth of the problems in each library to form the 200 problem set, whose formation is shown in Table~\ref{tab:problem_count_human_evaluation}.
API is defined as a function call, a method call or an attribute getter and setter in this case.
For retrieved knowledge, we follow DocPrompting~\cite{zhou2023docprompting} to take problem descriptions as queries and their provided document pool as the target sets to perform the retrieval.
We use their pretrained CodeT5 retriever and retrieve $k=5$ knowledge items from the target pool.
For irrelevant documents, we filter them for both retrieved documents and generated ones, so the final number of documents for each problem may be smaller than $k$.
After retrieval, we follow \citet{zhao2021calibrate} to put items with higher scores near the generation position to ensure the best generation result.

\begin{table}[htbp]
  \centering
  \caption{Problem counts for each library in the selected set.}
    \begin{tabular}{lccccccc}
    \toprule
    \textbf{Library} & \textbf{Tensorflow} & \textbf{Pytorch} & \textbf{Numpy} & \textbf{Matplotlib} & \textbf{Pandas} & \textbf{Sklearn} & \textbf{Scipy} \\
    \midrule
    \#Problems & 10    & 15    & 40    & 37    & 44    & 28    & 26 \\
    \bottomrule
    \end{tabular}%
  \label{tab:problem_count_human_evaluation}%
\end{table}%

\section{Prompt for the First-Stage of \autoknow~in Each task}
\label{sec:each_task_gen_prompt}
\begin{figure*}[htbp]
\centering
  \includegraphics[width=\linewidth]{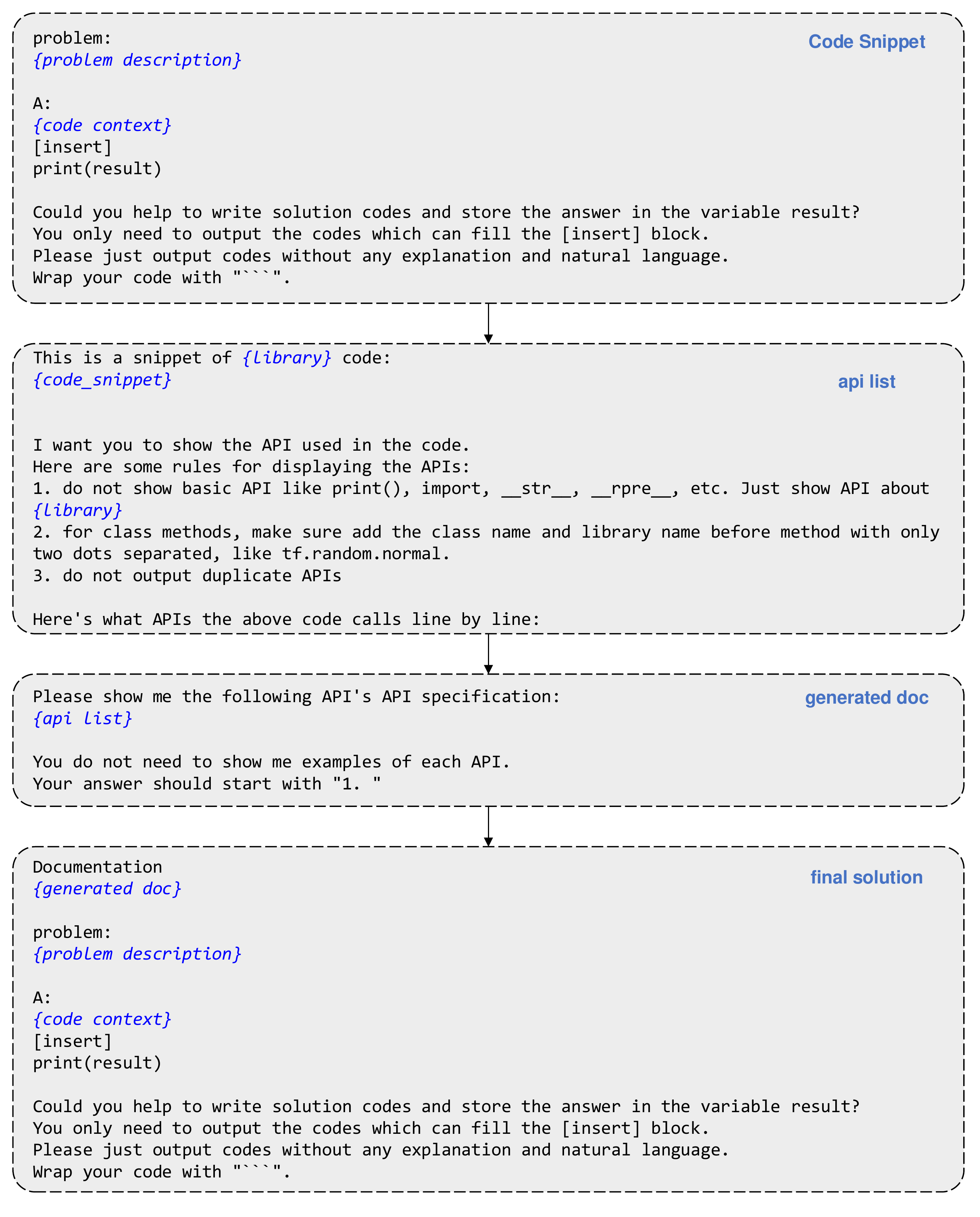} 
  \caption{Prompt for the first step of \autoknow~in the DS-1000 dataset.}
 \label{fig:prompt_ds1000_gen}
\end{figure*}

\begin{figure*}[htbp]
\centering
  \includegraphics[width=\linewidth]{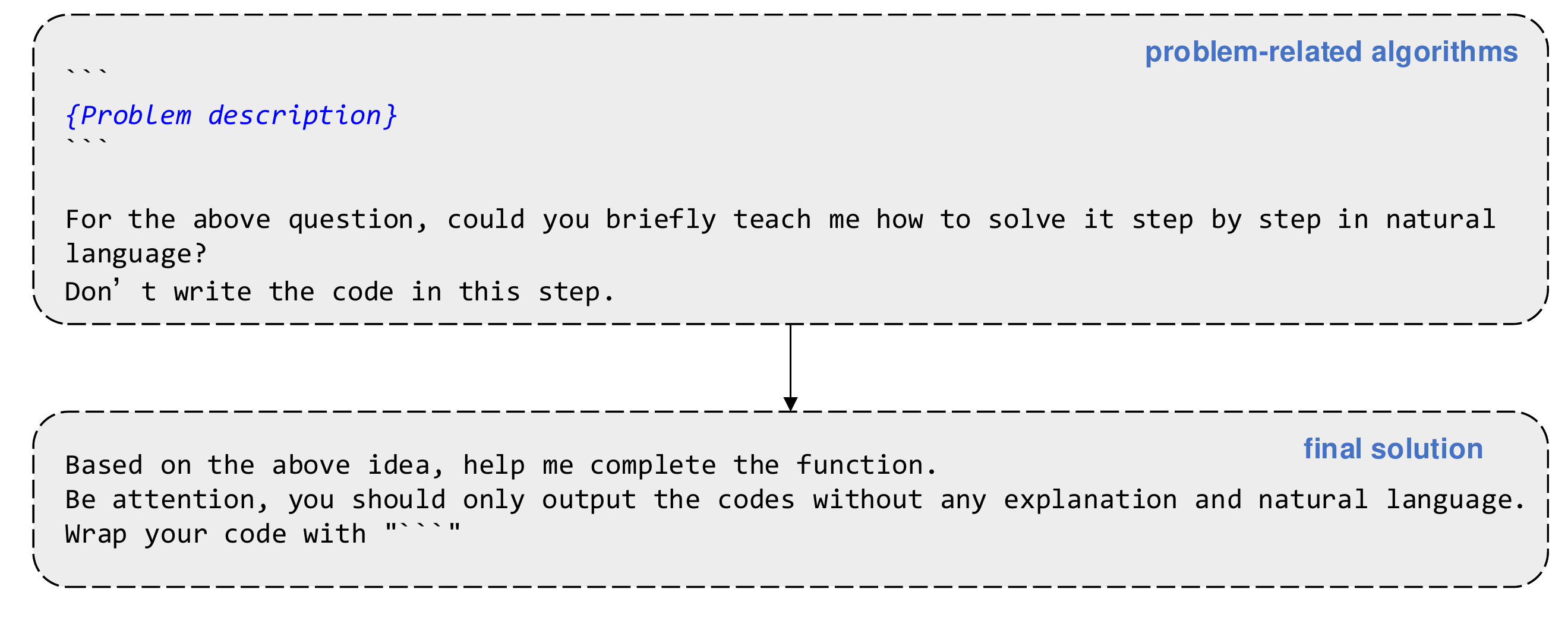} 
  \caption{Prompt for the first step of \autoknow~in the HumanEval dataset.}
 \label{fig:prompt_humaneval_gen}
\end{figure*}

\begin{figure*}[htbp]
\centering
  \includegraphics[width=\linewidth]{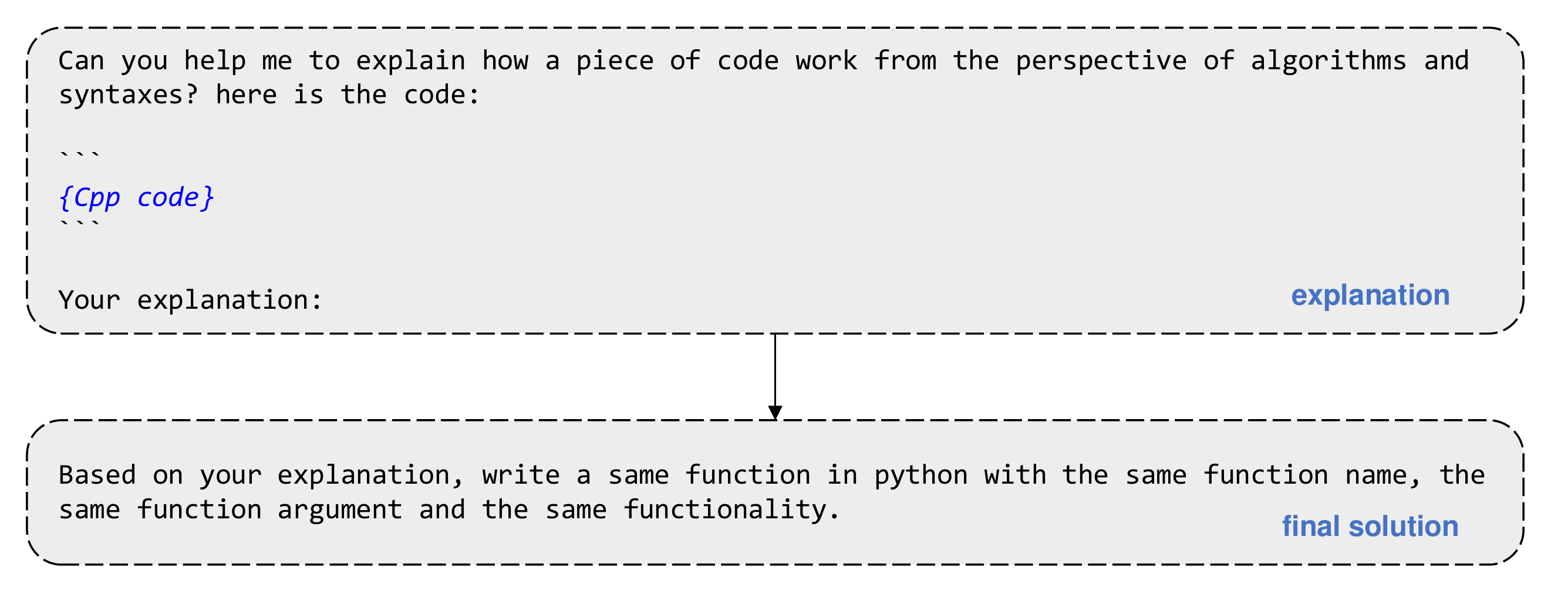} 
  \caption{Prompt for the first step of \autoknow~in the TransCoder dataset.}
 \label{fig:prompt_transcoder_gen}
\end{figure*}

We show the detailed prompt for the first step of \autoknow~used in each task below.
We show prompts for DS-1000, HumanEval and TransCoder in Figure~\ref{fig:prompt_ds1000_gen}, Figure~\ref{fig:prompt_humaneval_gen}, 
 and Figure~\ref{fig:prompt_transcoder_gen}, respectively.

\section{Prompt for Self-Refinement in Each task}
\label{sec:each_task_refine_prompt}
\begin{figure*}[htbp]
\centering
  \includegraphics[width=\linewidth]{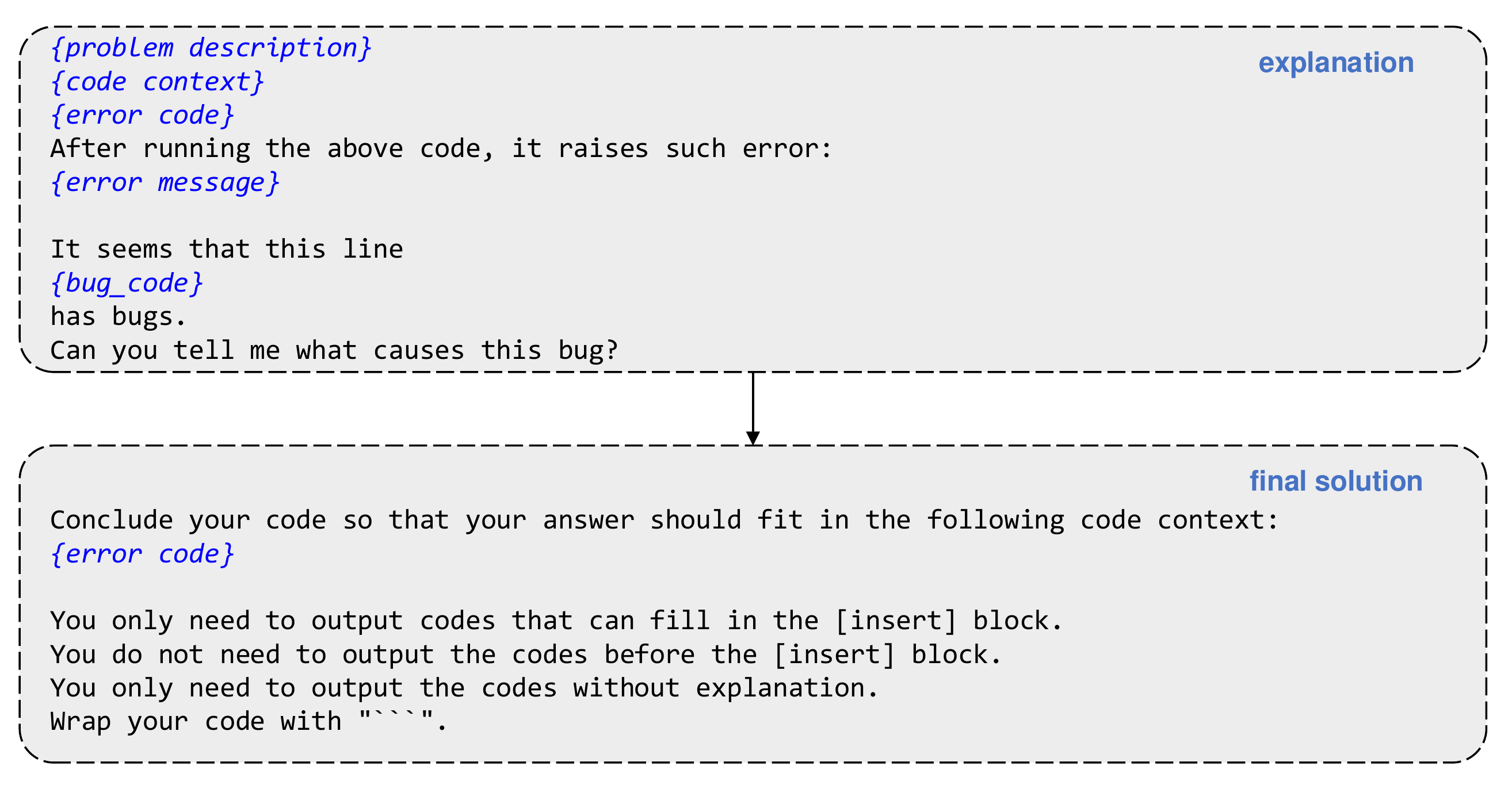} 
  \caption{Prompt for self-refinement in the DS-1000 dataset.}
 \label{fig:prompt_ds1000_revise}
\end{figure*}

\begin{figure*}[htbp]
\centering
  \includegraphics[width=\linewidth]{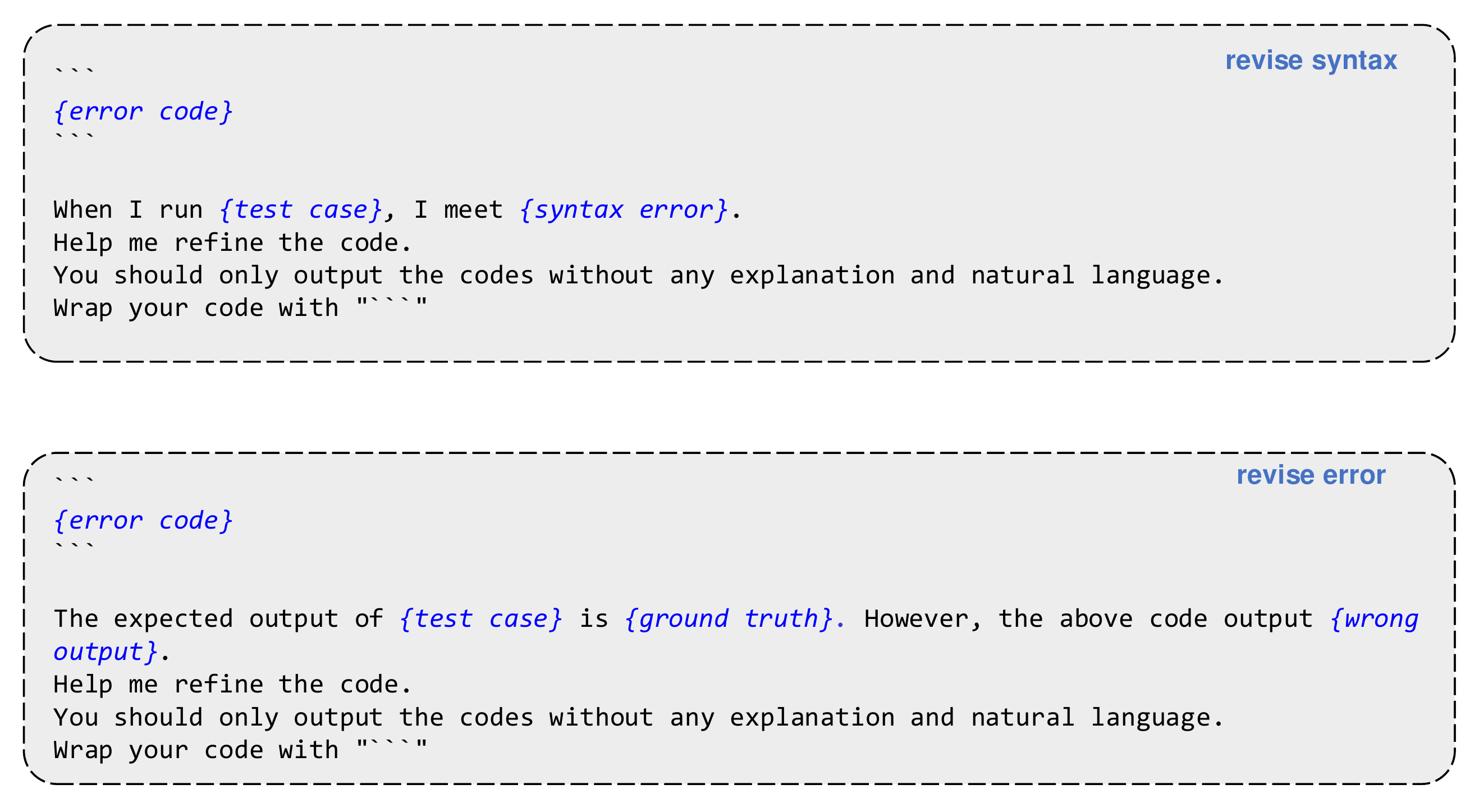} 
  \caption{Prompt for self-refinement in the HumanEval dataset.}
 \label{fig:prompt_humaneval_revise}
\end{figure*}

\begin{figure*}[htbp]
\centering
  \includegraphics[width=\linewidth]{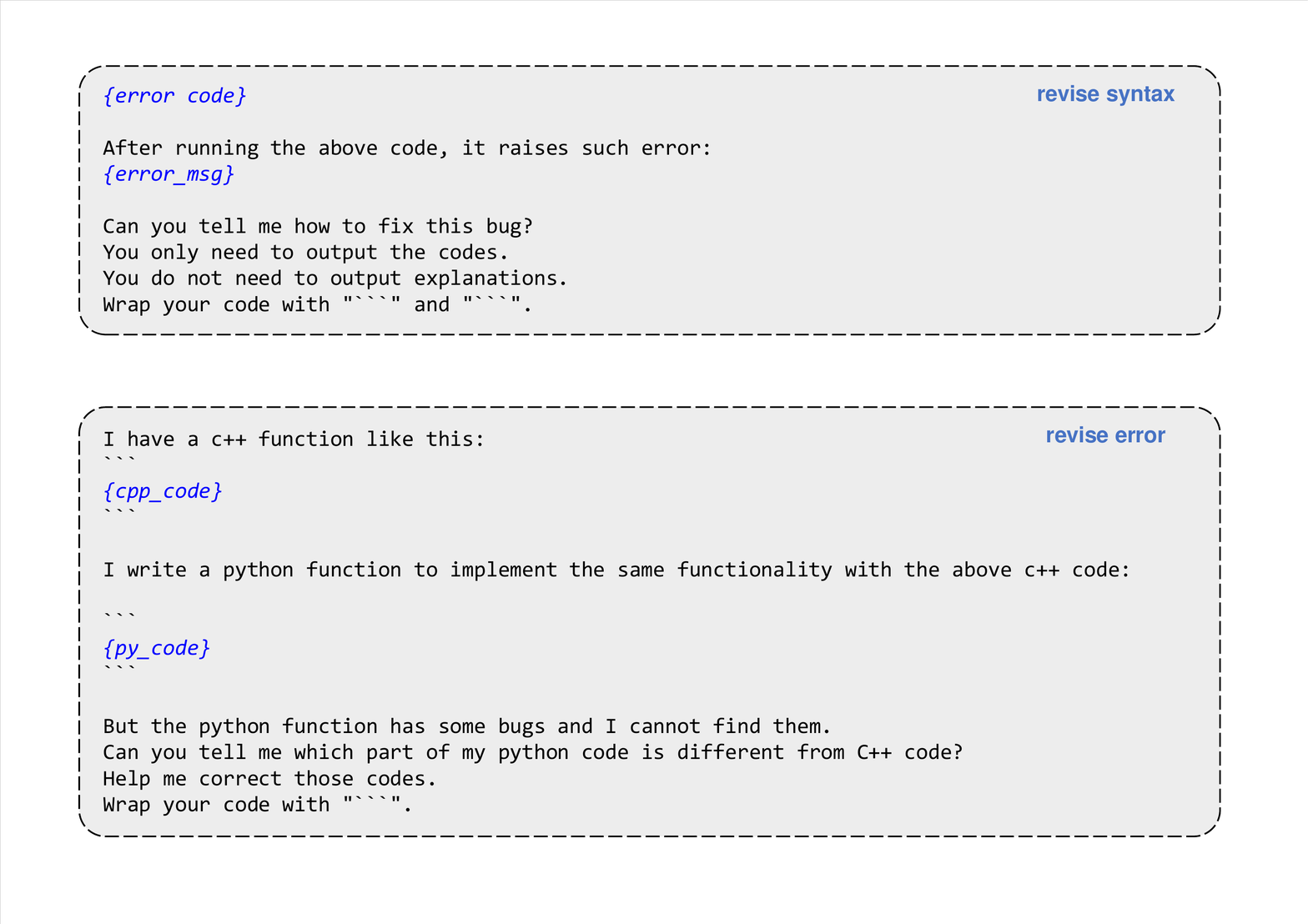} 
  \caption{Prompt for self-refinement in the TransCoder dataset.}
 \label{fig:prompt_transcoder_revise}
\end{figure*}

We show the detailed prompt for the self-refinement module used in each task below.
We show prompts for DS-1000, HumanEval and TransCoder in Figure~\ref{fig:prompt_ds1000_revise}, Figure~\ref{fig:prompt_humaneval_revise}, and Figure~\ref{fig:prompt_transcoder_revise}, respectively.

\end{document}